\documentclass[10pt,twocolumn,letterpaper]{article}

\usepackage[accsupp]{axessibility}
\usepackage{wacv}
\usepackage{times}
\usepackage{epsfig}
\usepackage{graphicx}
\usepackage{amsmath}
\usepackage{amssymb}
\usepackage{multirow}
\usepackage{soul}
\usepackage{bm}
\usepackage{enumitem}
\usepackage{booktabs}
\usepackage{pifont}% http://ctan.org/pkg/pifont
\newcommand{\cmark}{\ding{51}}%
\newcommand{\xmark}{\ding{55}}%

\def\wacvPaperID{1096} % Enter the WACV Paper ID here

\wacvfinalcopy % *** Uncomment this line for the final submission

\ifwacvfinal
\fi

\ifwacvfinal
\usepackage[breaklinks=true,bookmarks=false]{hyperref}
\else
\usepackage[pagebackref=true,breaklinks=true,colorlinks,bookmarks=false]{hyperref}
\fi

\ifwacvfinal
\else
\fi

\begin{document}

%%%%%%%%% TITLE
\title{YOLO-ReT: Towards High Accuracy Real-time Object Detection on Edge GPUs}

\author{Prakhar Ganesh$^{1*}$, Yao Chen$^{1}$\thanks{Both authors contributed equally to this research.} , Yin Yang$^2$, Deming Chen$^3$, Marianne Winslett$^3$ \\
$^1$Advanced Digital Sciences Center, Singapore \\
$^2$College of Science and Engineering, Hamad Bin Khalifa University, Qatar \\
$^3$University of Illinois at Urbana-Champaign, USA \\
{\tt\small \{prakhar.g, yao.chen\}@adsc-create.edu.sg, yyang@hbku.edu.qa, \{dchen, winslett\}@illinois.edu}
% \and
}

\maketitle
\thispagestyle{empty}

%%%%%%%%% ABSTRACT
\begin{abstract}
Performance of object detection models has been growing rapidly on two major fronts, model accuracy and efficiency. However, in order to map deep neural network (DNN) based object detection models to edge devices, one typically needs to compress such models significantly, thus compromising the model accuracy. In this paper, we propose a novel edge GPU friendly module for multi-scale feature interaction by exploiting missing combinatorial connections between various feature scales in existing state-of-the-art methods. Additionally, we propose a novel transfer learning backbone adoption inspired by the changing translational information flow across various tasks, designed to complement our feature interaction module and together improve both accuracy as well as execution speed on various edge GPU devices available in the market. For instance, YOLO-ReT with MobileNetV2$\times$0.75 backbone runs real-time on Jetson Nano, and achieves 68.75 mAP on Pascal VOC and 34.91 mAP on COCO, beating its peers by 3.05 mAP and 0.91 mAP respectively, while executing faster by 3.05 FPS. Furthermore, introducing our multi-scale feature interaction module in YOLOv4-tiny and YOLOv4-tiny (3l) improves their performance to 41.5 and 48.1 mAP respectively on COCO, outperforming the original versions by 1.3 and 0.9 mAP.
% \textcolor{red}{(Open source.)}
\end{abstract}

\section{Introduction}
\label{sec:intro}

Object detection is one of the most fundamental problems in computer vision, which does both localisation and classification of various objects in an image or a video.
Due to its foundational nature, object detection has numerous applications across various domains like unmanned autonomous vehicle (UAV) \cite{simhambhatla2019self}, medical imaging \cite{hafiz2020survey}, identity verification \cite{shi2019docface+}, robot navigation \cite{zhang2019skynet, shang2019survey}, sports analysis \cite{kamble2019ball}, \textit{etc.} and has been adapted into more complex problem statements like object tracking \cite{yao2019video}, action recognition \cite{beddiar2020vision}, face recognition \cite{loy2020face}, \textit{etc.} 
Many of these applications require highly accurate real-time feedback.

Object detection has seen major advancements in recent years both in accuracy as well as efficiency due to the adoption of deep learning algorithms \cite{chahal2018survey, liu2020deep}.
A majority of research in efficient object detection is dominated by single-stage object detection models that do both localisation and classification in the same network \cite{bochkovskiy2020yolov4, fang2019tinier, tan2020efficientdet}, and focuses on real-time execution using Desktop GPUs, like YOLOv4 \cite{bochkovskiy2020yolov4}, EfficientDet \cite{tan2020efficientdet}, ScaledYOLOv4 \cite{wang2021scaled}, \textit{etc.} However the execution speed of such models (measured in frames per seconds or FPS) drops dramatically on edge devices due to their high computational requirement \cite{mlfpgaiot}.

Real-time object detection is a vital functionality for modern end-node IoT devices, which due to their disconnection to a centrally located computing platform, do not pack large amounts of computing power and face stricter power supply and latency constraints.
However, bringing the power of computing directly to such devices where the data is being collected or consumed, can help us save the data transfer and high computational costs, and improve on response time and other bandwidth issues that would arise in case of an off-location computing. While most of the recent work in the field of lightweight computer vision focuses on reducing the model size and computations \cite{du2020spinenet, howard2019searching, tan2019efficientnet}, it is important to note that these improvements do not directly translate into a faster model. In fact, using these metrics of comparison without executing the model on target device can lead to sub-optimal design \cite{ma2018shufflenet,rakhimov2021making}.

Multi-scale feature interaction is at the heart of modern object detection models \cite{lin2017feature, liu2018path, tan2020efficientdet}. However, with the increasing complexity of these feature interaction modules, the trade-off between efficiency and accuracy is saturating (see Table \ref{tab:raw}), leaving the need for an innovative feature interaction method. Since most existing methods focus on some combination of the top-down or bottom-up approach to feature collection, these paths leave out a number of possible inter-scale interactions between non-adjacent feature scales that can significantly improve feature refinement for further processing. Additionally, existing feature interaction methods are constrained by the number of output scales, thus missing out on important low-level features.

The direct adoption of transfer learning backbone from classification to detection has been a topic of debate for a long time \cite{zhuang2020comprehensive}, with a number of research papers even creating their own backbones designed and trained directly on object detection datasets \cite{wong2019yolo, xiong2020mobiledets}. The increasingly task-specific nature of later layers in a backbone has also been studied extensively \cite{yosinski2014transferable}, and thus directly using pre-trained backbone during transfer learning is clearly not the best adoption of expert knowledge available at our disposal. A more effective backbone adoption is required to achieve the best trade-off between accuracy and efficiency. 

In this work, we design a raw feature collection and redistribution module along with an improved truncated backbone adoption during transfer learning that is compatible with various feature extraction backbones and detection heads, and improves both model's execution speed as well as accuracy on edge GPUs.
Our contributions include:

\begin{itemize}[leftmargin=*]
\itemsep0em
\item A lightweight raw feature collection and redistribution (RFCR) module that efficiently combines multi-scale features, compatible with various backbones and detection heads. Additionally, the feature collection of our RFCR module is independent of the number of output scales in the detection head, facilitating better feature interaction.
\item An extensive experimental analysis of the importance of individual transfer learning layers, together with a truncation method for improved model efficiency. Our truncation and RFCR module complement each other, allowing us to create faster and more accurate detection models.
\item An in-depth ablation study with on-device execution latency experiments for edge GPUs, instead of other indirect metrics like MFLOPs or model size, thus providing an accurate comparison of various competing designs.
\end{itemize}
\section{Background}
\label{sec:related}

Convolutional Neural Networks (CNNs) in the last decade have made several advancements in the direction of lightweight components which benefits both the feature extraction backbone as well as the head of an object detection model. We elaborate the details in the following.

\subsection{Single-stage Object Detection}

A modern single-stage object detection model comprises of two components, a feature extractor usually pre-trained on ImageNet \cite{russakovsky2015imagenet} and an object detection head responsible for the final output. While CNNs are the go-to choice for feature extraction models, there does exist some work on the exploration of other forms of feature extractors, e.g., extreme learning machines (ELM) \cite{yin2020faster}, motion probability maps \cite{shaifee2017fast}, \textit{etc.}.
Single-stage object detection models can be further divided based on the detection head they use into anchor-based or anchor-free models. Heatmap-based detection models like CornerNet \cite{law2018cornernet}, CenterNet \cite{duan2019centernet}, \textit{etc.} are common examples of anchor-free models. However, these models require computationally expensive backbones \cite{newell2016stacked} as they rely on keeping high resolution information of the input image intact. Anchor-based detection models on the other hand are the lighter alternatives. For example, YOLOv3 detection head \cite{redmon2018yolov3} is one of the most commonly used detection head for edge devices, and allows easy integration of lightweight backbones \cite{fang2019tinier, wong2019yolo, yolofastest, bochkovskiy2020yolov4, wang2021scaled}.

\subsection{Building Blocks}
\label{subsec:blocks}

A large section of research in real-time object detection models have been devoted to improving the basic building blocks of CNNs.
The traditional CNN layers contain a large number of parameters as well as computations, forcing most such real-time detection models to be significantly shallow networks \cite{fang2019tinier, redmon2018yolov3}.
% \textcolor{blue}{I feel the logic here is a bit improper: shallow networks are still using conventional Conv layers. So, here, actually 3 techniques are presented, shallow nets, decoupling and 1x1}. 
Decoupling 2D convolution into depthwise separable and pointwise ($1\times1$) convolutions is a common technique to make networks lighter \cite{yolofastest, huang2018yolo, tan2020efficientdet, zhao2020mixed}.
Further reducing the number of channels using $1\times1$ convolutions before applying the intended convolution gave birth to the idea of fire modules \cite{iandola2016squeezenet} and have been adapted in various lightweight detection models \cite{fang2019tinier, law2019cornernet, li2020real, womg2018tiny, wu2017squeezedet}.

However, using multiple consecutive pointwise convolutions to reduce the computational cost of the information flow infringes on an essential rule of designing fast deep learning models, i.e., network fragmentation \cite{ma2018shufflenet}.
Network fragmentation is a phenomena in which a heavier operation is fragmented into multiple lightweight operations, and significantly hurts the model's execution speed as it interferes with its internal degree of parallelism \cite{ma2018shufflenet}.
For example, MobileDets \cite{xiong2020mobiledets} discovered that grouped pointwise convolutions are not well executed on GPU devices, while ShuffleNetV2 \cite{ma2018shufflenet} found that pointwise convolutions are fastest when number of input and output channels are the same.

The final feature extraction backbone is formed by combining one or more of the building blocks mentioned above. A number of works have even utilised Neural Architecture Search (NAS) to build their own backbones and detection models \cite{xiong2020mobiledets, wong2019yolo}. 
However, these models miss out on transfer learning information present in other pre-trained backbones \cite{li2020mspnet, li2018tiny}.
On the other hand, backbones pre-trained on existing datasets might contain classification task specific features \cite{li2018detnet, xiong2020mobiledets}, which can add an unnecessary burden of feature calculation. Thus, an efficient adaptation of a pre-trained backbone from classification to object detection also plays a major role in the model's final performance.

\subsection{Multi-scale Feature Fusion}
\label{sebsec:related_fusion}

Multi-scale feature interaction is a vital part of the object detection head, both in single-stage as well as two-stage object detection models. 
Existing methods of feature interaction take some combination of either the top-down or bottom-up approach for the flow of information across multi-scale features \cite{hurtik2020poly, lin2017feature, liu2018path, mao2019mini, qin2019thundernet, tan2020efficientdet, yoo2019scarfnet, zhao2019m2det}. Feature Pyramid Networks (FPN) \cite{lin2017feature} were the first to create a top-down path from high-level feature scales towards low-level feature scales, with the purpose of using well processed deeper features to help improve the accuracy of detection layers using shallower features. Path Augmentation Networks (PANet) \cite{liu2018path} took it a step further and showed that an additional bottom-up path can help further improve the detection accuracy of high-level features.

Building on the success of FPN and PANet, NAS-FPN \cite{chen2020mnasfpn, ghiasi2019fpn} attempted to find the optimal paths of information flow between various multi-scale features. Since such architecture search based models are designed specifically for certain datasets and backbone networks, it is difficult to generalise them to a wider range of applications. However, these searches reveal interesting trends that can help us learn more about the inherent requirements of such models. NAS-FPN designs revealed the presence of direct connections between various feature scales not adjacent to each other, showing that the flow of information only through adjacent scales might become convoluted and warrants the need of such shortcut connections. Similarly, NAS-FPN also revealed the importance of repeatedly following the top-down and bottom-up path that was later adopted by BiFPN \cite{tan2020efficientdet} to further improve model accuracy.

Not only the path taken to combine multi-scale features together, but a lot of work have also been done on how various features are combined. While most existing work simply concatenates feature maps from multiple scales together, weighted or attention-based fusion of features have also been proposed \cite{li2020mspnet, tan2020efficientdet} to better highlight more important feature scales.
Another aspect of fusing features is bringing them to a common scale. Simpler solutions for this includes upsampling or downsampling one of the feature scales to match the other. However, this can entail a local positional mismatch between various scales, and thus multiple ways have also been explored to process the features before and after fusion to facilitate better flow of information across various scales \cite{tang2020lightdet, zhang2019skynet, dai2021attentional}.
\section{Proposed Solution}
\label{sec:solution}

\begin{figure*}
\centering
\includegraphics[width = 0.9\textwidth]{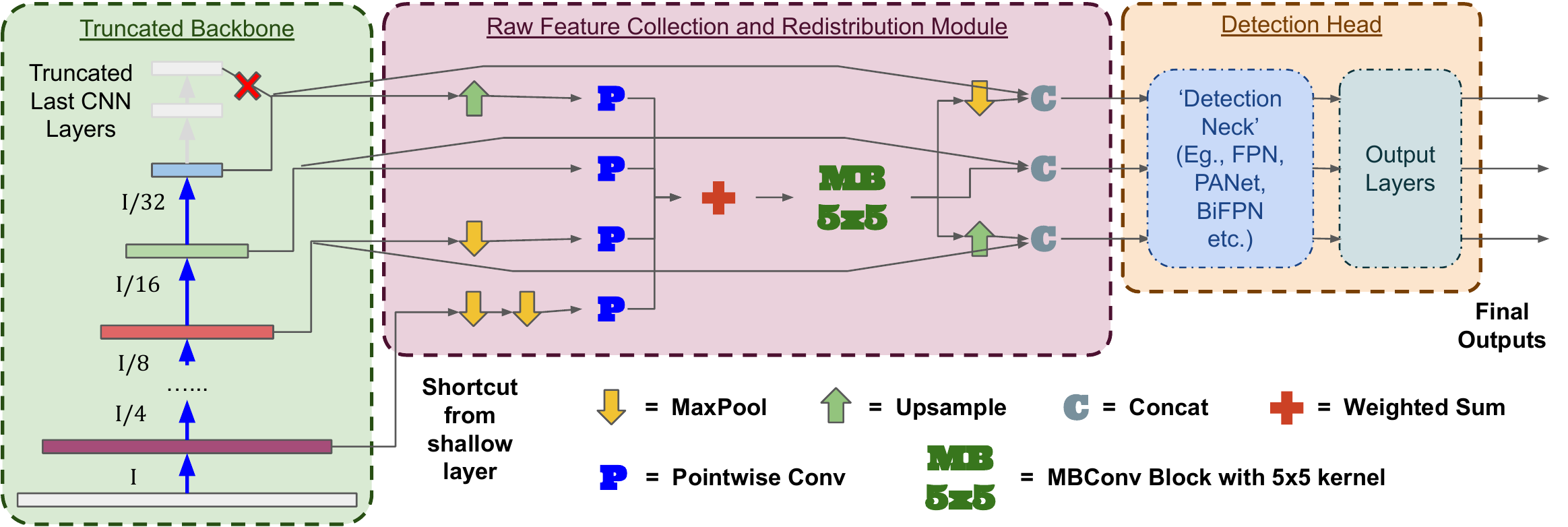}
\caption{Complete Architecture of YOLO-ReT.}
\label{fig:complete}
\end{figure*}

In this section, we introduce a class of object detection models, YOLO-ReT, which uses our RFCR module and transfer learning inspired backbone truncation to improve both accuracy and efficiency on edge GPUs.

\subsection{Raw Feature Collection and Redistribution}
\label{subsec:rfcr}
% \textcolor{red}{1. Figure 1: add in neck and head.
% 2. Add in the scale related presentation.}
We first introduce our raw feature collection and redistribution (RFCR) module.
Following our discussion in Section \ref{sebsec:related_fusion}, we look to strengthen the raw features provided by the backbone using an improved feature interaction network to increase the detection accuracy, without causing any significant harm to the execution speed. While we focus specifically on detection in this paper, our RFCR module can be generalised to feature interaction for similar tasks.

Existing methods of multi-scale feature interaction can be broken down into some combination of the top-down and bottom-up approaches which focuses on only two adjacent feature scales at a time. This misses out on a large number of possible combinatorial pairs and makes the propagation of information between distant feature scales inefficient \cite{chen2020mnasfpn, ghiasi2019fpn}. Furthermore, when repeatedly using the top-down and bottom-up paths like in BiFPN \cite{tan2020efficientdet}, e.g., moving from BiFPNx2 to BiFPNx3, the detection accuracy of the model starts to saturate (see Table \ref{tab:raw} for details).
% \textcolor{red}{which motivates the need for a better feature interaction method}.

Here, inspired by non-adjacent feature scale connection in NAS-FPN \cite{chen2020mnasfpn, ghiasi2019fpn}, we propose a lightweight feature collection and redistribution module which fuses raw multi-scale features from the backbone together and then redistributes it back to each feature scale. Thus feature maps from each scale now contain direct connections from all the other scales. Such a layer does not involve any heavy computations or parameters, however allows a direct link between every pair of feature scales, as shown in the Figure \ref{fig:complete}. It should be noted that our RFCR module cannot replace the meticulousness that other feature aggregation methods provide, but instead we aim to provide an extremely lightweight feature processing before passing them to other multi-scale feature fusion methods, providing orthogonal improvements in accuracy.

Additionally, our module design allows us independence from the number of output scales in the detection head, as there are no constraints between the number of input and output features to our RFCR module. For example, despite YOLOv3 detection head having 3 output scales, we can use four different backbone features (3 features same as the output scales, with a fourth shallower feature 'shortcut') during feature collection stage, allowing us to utilize more fine-grained low-level features to improve model performance \cite{wang2019learning}. Similarly, even for detection heads with only 2 output scales like in YOLOv4-tiny \cite{wang2021scaled}, detection features are enriched by the multiple low-level features with the adoption of our RFCR module (see Section \ref{sec:yolov4}).

As discussed in Section \ref{sebsec:related_fusion}, the manner of feature fusion is as important as the aggregation path.
In order to keep the additional latency overhead to a minimum, we pass the raw features during collection through a single 1x1 convolution, and use a simple weighted sum to fuse features together. We pass the fused feature map through a mobilenet convolution block (MBConv), which is then redistributed back to various scales. Such a design allows us to keep the network fragmentation to a minimum, since our RFCR module can be represented with just four layers, a 1x1 convolution, a weighted sum and two layers in the MBConv block, along with upsampling and downsampling layers as required.
The parallel collection and redistribution of features can also be easily optimised for faster execution.

When fusing features from different scales, naive upsampling and downsampling can cause inconsistent semantics and local positional mismatch \cite{dai2021attentional}. Thus, we propose increasing the receptive field of the feature fusion layer by using a 5x5 kernel instead of the conventional 3x3 or 1x1, to help improve the detection accuracy of the model with negligible affect on its execution latency.
We found increasing the kernel to 7x7 did not benefit the performance further.

\subsection{Backbone Truncation}

\begin{figure}
\centering
\includegraphics[width = 0.45\textwidth]{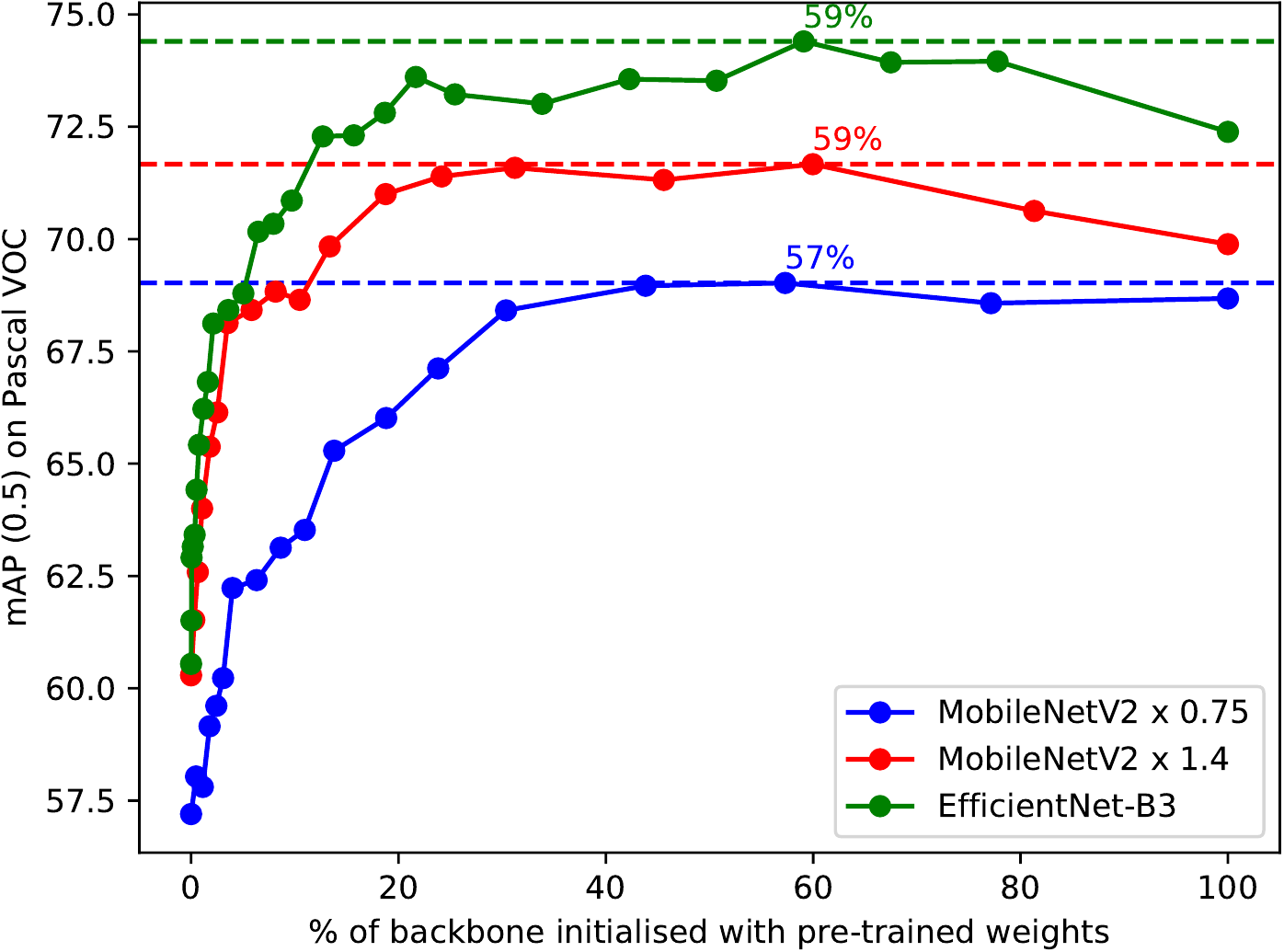}
\caption{Transfer Learning Curve.}
\label{fig:transfer}
\end{figure}

Most state-of-the-art lightweight image classification models \cite{ma2018shufflenet, sandler2018mobilenetv2, tan2019efficientnet, zhang2018shufflenet} attempt to keep the number of channels to a minimum by gradually increasing them after every few convolution blocks. However, towards the end, even these models start rapidly expanding the number of channels after every block in an attempt to represent features more clearly before the final fully connected layer \cite{howard2019searching, tan2019mnasnet}.
Not only are these last CNN layers the most computationally expensive and heaviest part of the backbone (see the spacing between datapoints in Figure \ref{fig:transfer}), but since they are used to create a better representation for the final classification, these layers mainly contain task specific features. 

The importance of transfer learning from classification models has been questioned before, with certain papers even designing specialised backbones for detection \cite{xiong2020mobiledets, wong2019yolo}.
This is based on the intuition that the translational information flow (i.e. across height and width of the image) through consecutive CNN layers varies across tasks. For example, classification models do not preserve spatial information and might accumulate to a spatially coarse feature. On the other hand, detection models attempt to keep the spatial information intact, required for a fine-grained detection output. 
We identify that the transfer learning capabilities of the initial layers of the feature extraction backbone are quite vital, and it is the last layers that do not provide critical information for the detection/recognition.

We test the importance of individual backbone convolution layers with a detailed analysis of the transfer learning capabilities of various feature extraction backbones, complete with PANet \cite{liu2018path} feature aggregation path and YOLOv3 \cite{redmon2018yolov3} detection head. 
We experiment with three commonly used backbones, MobilenetV2 ($\times$0.75 and $\times$1.4) and EfficientNet-B3 for our experiments (see Section \ref{sec:eval_setup} for more details) and divide the backbone into various blocks, which in this case are MBConv blocks for MobileNetV2 and MBConvSE blocks for EfficientNet.
Next, we gradually increase, from shallower towards deeper, the number of blocks which are initialised using pre-trained weights from ImageNet dataset while the rest are initialised randomly similar to the detection head, and trained each individual model to convergence (see Section \ref{sec:eval_setup} for the training setup). The collected results can be seen in Figure \ref{fig:transfer}.

It can be noted from the figure, as we increase the portion of feature extraction backbone initialised with pre-trained weights, the model performance improves, emphasizing the importance of transfer learning. However, around the 60\% mark the performance starts to deteriorate and fluctuate. This shows that initializing the last layers of feature extractors with transfer learning weights from ImageNet actually damages the performance as compared to random initialization, possibly because of being stuck in a local minima due to the task-specific nature of these layers.

Since these last layers hold no transfer learning importance, they can be analysed purely from an architecture viewpoint. As can be seen in Figure \ref{fig:transfer}, these last 2 or 3 layers contain over 40\% of the weights due to an extreme expansion of number of channels not relevant for object detection.
Thus, we propose using a truncated version of various feature extraction backbones for the final object detection model, presenting it as a better alternative than reducing the scaling factor. We use the results from Figure \ref{fig:transfer} to find the point of truncation, i.e., we truncate the last two blocks from both MobileNetV2 versions, and the last three blocks from EfficientNet when we adopt them as backbones.

\section{Evaluations}
\label{sec:eval}
We evaluate our proposed solution using mean average precision \cite{lin2014microsoft} on multiple datasets as well as FPS achieved on various edge GPU devices.
A thorough ablation study of our proposed RFCR module and backbone truncation and comparison of our model's performance against various real-time object detection models is conducted.

\subsection{Evaluation Setups}
\label{sec:eval_setup}

\noindent \textbf{Feature Extraction Backbone.}
The balance of execution speed and accuracy of the feature extraction backbone is crucial to the final performance of the detection model due to the transfer learning capability it offers~\cite{bochkovskiy2020yolov4, duan2019centernet, tan2020efficientdet}. As discussed in the Section \ref{subsec:blocks}, the device friendliness together with the theoretical computation and model size are all important factors that affect the model's execution speed. Instead of theoretically modeling the impact of these factors, we directly collect the FPS of various feature extraction backbones on Jetson devices in Table~\ref{tab:backbone}.
Based on the table, we focus on three commonly used lightweight backbones, MobilenetV2 ($\times$0.75), MobilenetV2 ($\times$1.4) and EfficientNet-B3 for our ablation study.

\begin{table}
\scriptsize
% \footnotesize
\begin{center}
\setlength\tabcolsep{2.8pt}
\begin{tabular}{c|c|c|ccc|c|c}
\toprule[2pt]
\multirow{2}{*}{\textbf{Backbone}} & \multirow{2}{*}{\textbf{Depth}} & \multirow{2}{*}{\textbf{Acc.}} & \multicolumn{3}{|c|}{\textbf{FPS}} & \textbf{Size} & \multirow{2}{*}{\textbf{Comp.}} \\
% \cline{4-6}
 & & & \textbf{Nano} & \textbf{NX} & \textbf{AGX} & \textbf{(MB)} & \\
\hline
\hline

ResNet50 \cite{he2016deep} & 177 & 74.9\% & 33.44 & 102.14 & 166.17 & 97.8 & 3989 \\
% MblNetV2 x 0.75 \cite{sandler2018mobilenetv2} & 157 & 69.8\% & 78.57 & 183.08 & 262.74 & 10.2 & 188 \\
MblNetV2 x 1.4 \cite{sandler2018mobilenetv2} & 157 & 74.7\% & 47.64 & \textbf{129.96} & \textbf{188.64} & 23.5 & 588 \\
DarkNet19 \cite{redmon2017yolo9000} & 62 & 72.9\% & \textbf{48.37} & 115.07 & 187.25 & 79.5 & 2764 \\
DarkNet53 \cite{redmon2017yolo9000} & 187 & 77.2\% & 20.81 & 71.53 & 142.99 & 158 & 7172 \\
CSPDarkNet53 \cite{wang2020cspnet} & 418 & 77.2\% & 20.60 & 82.49 & 153.08 & 109 & 5038 \\
EfficientNet-B0 \cite{tan2019efficientnet} & 250 & 77.1\% & 36.91 & 112.09 & 177.97 & \textbf{20.4} & \textbf{396} \\
EfficientNet-B3 \cite{tan2019efficientnet} & 407 & \textbf{81.6\%} & 17.97 & 74.48 & 134.94 & 47.1 & 1007 \\

\bottomrule[2pt]

\end{tabular}
\end{center}
% \vspace{-6pt}
\caption{Backbone accuracy (ImageNet Top-1) and FPS}\label{tab:backbone}
% \vspace{-12pt}
\end{table}

\noindent \textbf{Lightweight Detection Layers.}
We use feature aggregation path proposed by PANet \cite{liu2018path} and YOLOv3 detection head \cite{redmon2018yolov3}. Instead of traditional convolution blocks as in PANet, we chose single 1x1 pointwise convolution layers \cite{howard2017mobilenets} for all forms of connections between various feature levels.
All features are passed through a single MBConvSE block \cite{tan2019efficientnet} before every feature aggregation path. We also do all feature refining during feature aggregation, and again use only a single 1x1 pointwise layer for converting the final feature map into an object detection output.

\noindent \textbf{Dataset and Evaluation Metrics.}
We choose Pascal VOC \cite{everingham2010pascal} and COCO \cite{lin2014microsoft} datasets for their popularity for similar tasks.
Pascal VOC contains a total of 16,551 images, with 20 object classes and an average of 2.4 bounding boxes per image. COCO is a larger dataset that contains 117,264 images with 80 object classes and has an average of 7.4 bounding boxes per image.
We conduct all ablation studies on Pascal VOC, while the final comparison with other state of the art lightweight object detection models in literature are on both datasets. We use Pascal VOC 2007 and 2012 training dataset together for training and compare results on the 2012 test split. For COCO, we use the train and validation splits of COCO 2017 dataset.
For Pascal VOC, we use the default IoU threshold of 0.5, while for COCO we provide various detailed metrics as proposed by the dataset \cite{lin2014microsoft}. 
We also compare the runtime FPS of various models on Jetson Nano, Jetson Xavier NX and Jetson AGX Xavier.

\noindent \textbf{Training Details.}
We use two forms of data augmentation during training, (i) geometric augmentations like random crop, rotation, flip, resize etc., and (ii) photometric augmentations like HSV adjustment, brightness adjustment etc.
We use self-adversarial training \cite{bochkovskiy2020yolov4} along with a cosine learning rate decay \cite{loshchilov2016sgdr} for best results. 
For localization, we use GIoU loss \cite{rezatofighi2019generalized} rather than the other alternatives  (CIoU\cite{zheng2020distance}, MSE loss).
For training, we first freeze the layers which are initialised using transfer learning weights and train for 100 epochs with a starting learning rate of 0.001. Next, we unfreeze all the layers and fine-tune the model for 150 epochs on a smaller starting learning rate of 0.0001. 
We do all our ablation studies on 320x320 input resolution, but we also train 224x224 and 416x416 models for final comparison.

\noindent \textbf{Device Setup.}
For ablation study, all models together with the baselines are implemented on three Jetson devices for FPS calculation.
We use TensorRT based FP16 optimization to further speed up the execution. All models are executed with batch size = 1 as we are targeting real-time applications. We process 10,000 images and take the average execution time for FPS calculations. We also provide FPS values for TensorRT based INT8 optimizations, to provide a fair comparison against baseline models designed specifically for integer computations \cite{wong2019yolo}. 
It should be noted that Jetson Nano does not have tensor cores to support INT8 based optimization, and thus such models do not have any advantage over their floating point counter-parts on Jetson Nano. Additionally, the FPS values reported here might differ from the original reported values for various models, which can be attributed to TensorRT optimizations not performed, prediction box post-processing time not included or a bigger batch size used by certain works in literature.

\subsection{Ablation Study}

Following the steps to construct the final detection model, we first start with the backbone truncation and then integrate our RFCR module for final comparison.

\begin{figure*}
\centering
\includegraphics[width = 0.98\textwidth]{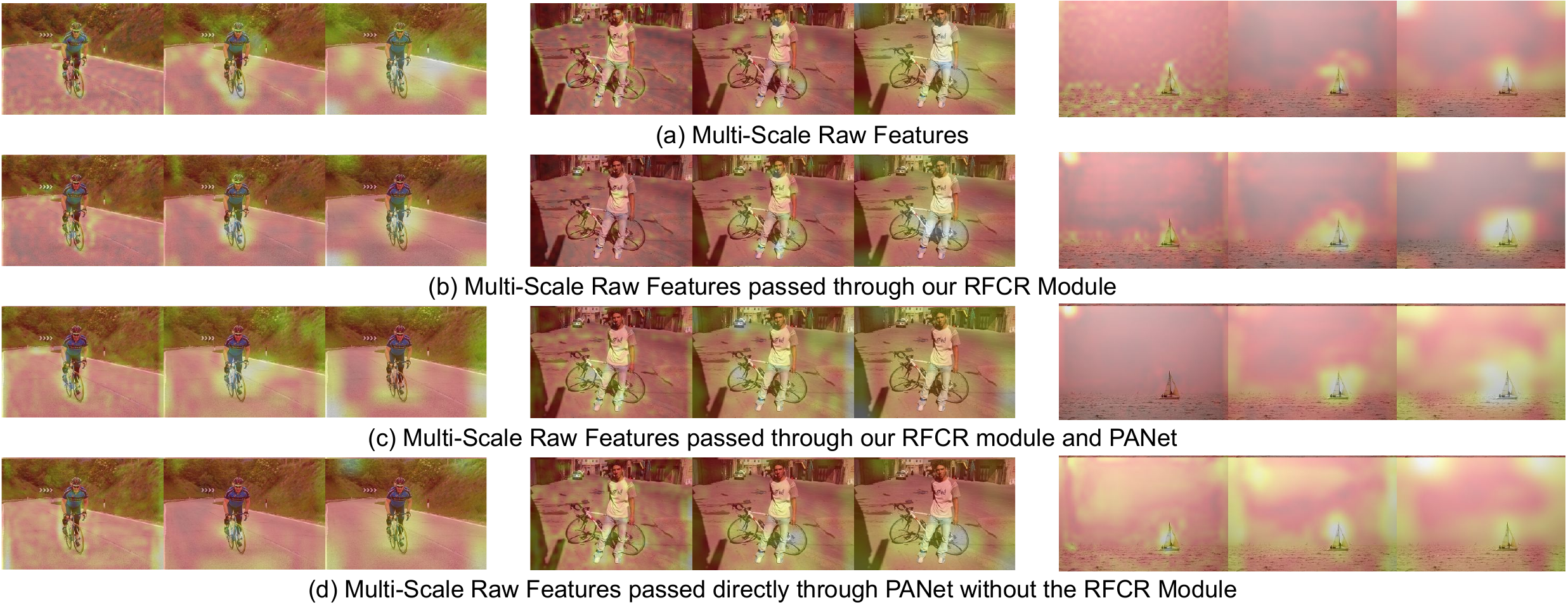}
\caption{Qualitative study with intermediate heatmaps. For all sets of three heatmaps, the original resolution of the feature maps from low-level to high-level features, i.e., left-to-right, is 40x40, 20x20 and 10x10 respectively, for a 320x320 input image.}
\label{fig:qualitative}
\end{figure*}

\subsubsection{Truncated feature extraction backbone}

\begin{table}
\scriptsize
% \footnotesize
\begin{center}
\setlength\tabcolsep{2.3pt}
\begin{tabular}{c|c|c|ccc|c|ccc}
\toprule[2pt]
 & & \multicolumn{4}{|c|}{\textbf{Complete Backbone}} & \multicolumn{4}{|c}{\textbf{Truncated Backbone}} \\
\cline{3-10}
 & \textbf{x} & \multirow{2}{*}{$\bm{AP^{50}}$} & \multicolumn{3}{|c|}{\textbf{FPS}} & \multirow{2}{*}{$\bm{AP^{50}}$} & \multicolumn{3}{|c}{\textbf{FPS}} \\
% \cline{4-6}
% \cline{8-10}
 & & & \textbf{Nano} & \textbf{NX} & \textbf{AGX} & & \textbf{Nano} & \textbf{NX} & \textbf{AGX} \\
\hline
\hline

 & 1.4 & 69.88 & 19.96 & 62.58 & 90.63 & 69.67 & 24.58 & 67.38 & 95.19 \\
\textbf{MblNetV2} & 1.0 & 69.40 & 24.85 & 67.92 & 93.20 & 68.87 & 29.32 & 73.87 & 98.67 \\
\textbf{(width = x)} & 0.75 & 68.67 & 28.16 & 73.25 & 99.75 & 66.58 & 34.02 & 77.67 & 103.77 \\
 & 0.5 & 63.94 & 35.18 & 81.05 & 117.52 & 61.27 & 39.97 & 86.13 & 124.27 \\
\hline

 & B3 & 72.05 & 9.69 & 45.72 & 73.15 & 72.28 & 11.24 & 49.72 & 78.61 \\
\textbf{Efficient-} & B2 & 71.84 & 12.44 & 50.86 & 80.08 & 71.92 & 13.45 & 55.61 & 82.47 \\
\textbf{-Net-x} & B1 & 71.67 & 13.34 & 52.31 & 82.19 & 71.59 & 14.62 & 57.52 & 86.22 \\
 & B0 & 71.24 & 18.27 & 60.97 & 95.96 & 70.98 & 19.08 & 64.34 & 99.95 \\

\bottomrule[2pt]

\end{tabular}
\end{center}
% \vspace{-6pt}
\caption{Comparing Width reduction vs Depth reduction}\label{tab:neck}
% \vspace{-12pt}
\end{table}

We compare two ways of compressing the MobileNetV2 and EfficientNet backbones, which are reducing the scaling factor (or width multiplier for MobileNetV2) and truncating the last parameter heavy layers, and collect the results in Table \ref{tab:neck}.
It can be noted that the truncated versions of EfficientNet are able to outperform their counterparts, both in terms of accuracy as well as FPS, emphasizing on the negative impact of classification task specific backbone features.

For MobileNetV2, when comparing models with similar FPS, the one with the truncated backbone performs better than the the one with smaller scaling factor. e.g., when comparing truncated backbone MobileNetV2x1.4 with complete backbone MobileNetV2x1.0, they both provide similar FPS while the former one provides 0.27 better mAP. 
This attributes to the fact that reducing the width multiplier reduces the number of channels uniformly across all the layers while truncating the backbone only removes the features from last layers.
This difference is exaggerated even more for lighter models on low power devices. 
For example, truncated backbone at width 0.75 for MobileNetV2 gives similar FPS to the complete backbone at width 0.5 (34.02 and 35.18 respectively on Jetson Nano), yet provides a 2.64 points improvement in mAP. Clearly, as we move towards real-time performing models, using a truncated backbone provides a more accurate and faster feature extraction network than the complete backbone.

\subsubsection{Raw feature collection and redistribution}

\begin{table}
\scriptsize
% \footnotesize
\begin{center}
\setlength\tabcolsep{1.6pt}
\begin{tabular}{c|c|c|ccc|c|ccc}
\toprule[2pt]
 & \textbf{Feature} & \multicolumn{4}{|c|}{\textbf{Without RFCR module}} & \multicolumn{4}{|c}{\textbf{With RFCR module}} \\
\cline{3-10}
\textbf{Backbone} & \textbf{Aggr.} & \multirow{2}{*}{$\bm{AP^{50}}$} & \multicolumn{3}{|c|}{\textbf{FPS}} & \multirow{2}{*}{$\bm{AP^{50}}$} & \multicolumn{3}{|c}{\textbf{FPS}} \\
% \cline{4-6}
% \cline{8-10}
 & \textbf{Path} & & \textbf{Nano} & \textbf{NX} & \textbf{AGX} & & \textbf{Nano} & \textbf{NX} & \textbf{AGX} \\
\hline
\hline

 & None & 59.92 & 38.24 & 80.17 & 126.42 & 63.97 & 36.93 & 78.38 & 119.41 \\
 & FPN & 64.15 & 37.82 & 78.42 & 111.02 & 66.11 & 36.33 & 76.95 & 101.94 \\
\textbf{MblNetV2} & PANet & 66.58 & 34.02 & 77.67 & 103.77 & 68.75 & 33.19 & 71.64 & 95.97 \\
\textbf{x 0.75} & BiFPNx1 & 66.29 & 34.81 & 78.04 & 103.47 & 66.65 & 33.78 & 72.17 & 95.70 \\
 & BiFPNx2 & 66.69 & 32.98 & 76.10 & 101.78 & 66.83 & 31.43 & 75.61 & 99.40 \\
 & BiFPNx3 & 66.78 & 31.54 & 74.45 & 100.37 & 66.90 & 30.72 & 73.68 & 98.27 \\
\hline

 & None & 68.15 & 28.21 & 72.97 & 110.24 & 69.26 & 26.50 & 71.81 & 106.43 \\
 & FPN & 69.02 & 26.06 & 69.69 & 103.18 & 69.95 & 24.19 & 66.23 & 98.73 \\
\textbf{MblNetV2} & PANet & 69.67 & 24.58 & 67.38 & 95.19 & 70.35 & 23.01 & 65.37 & 93.49 \\
\textbf{x 1.4} & BiFPNx1 & 69.50 & 25.26 & 67.42 & 95.55 & 70.14 & 23.60 & 65.79 & 93.71 \\
 & BiFPNx2 & 69.84 & 23.25 & 65.77 & 92.96 & 70.53 & 22.18 & 64.43 & 92.03 \\
& BiFPNx3 & 69.87 & 20.99 & 62.81 & 91.50 & 70.61 & 20.33 & 62.17 & 90.93 \\
\hline

 & None & 71.24 & 12.21 & 62.31 & 87.60 & 72.37 & 11.94 & 59.84 & 84.58 \\
 & FPN & 71.60 & 11.75 & 56.04 & 85.84 & 72.63 & 11.34 & 53.28 & 82.54 \\
\textbf{Efficient-} & PANet & 72.28 & 11.24 & 49.72 & 78.61 & 72.96 & 10.96 & 47.07 & 75.61 \\
\textbf{Net-B3} & BiFPNx1 & 72.07 & 12.12 & 47.70 & 79.28 & 72.80 & 11.71 & 45.02 & 75.39 \\
 & BiFPNx2 & 72.39 & 11.15 & 43.53 & 76.22 & 73.01 & 10.72 & 42.24 & 72.89 \\
 & BiFPNx3 & 72.51 & 9.92 & 39.91 & 72.55 & 73.08 & 9.38 & 38.25 & 67.99 \\
\bottomrule[2pt]

%  & None & 62.17 & 20.77 & -- & -- & 13.6 & 67.73 & 19.41 & -- & -- & 14.5 \\
%  & FPN & 68.43 & 18.98 & -- & -- & 14.0 & 70.34 & 18.87 & -- & -- & 15.5 \\
% \multirow{2}{*}{\textbf{EfficientNet-B0}} & PANet & 70.98 & 18.08 & -- & -- & 15.1 & 71.82 & 17.82 & -- & -- & 17.4 \\
%  & BiFPNx1 & -- & -- & -- & -- & -- & -- & -- & -- & -- & -- \\
%  & BiFPNx2 & -- & -- & -- & -- & -- & -- & -- & -- & -- & -- \\
%  & BiFPNx3 & -- & -- & -- & -- & -- & -- & -- & -- & -- & -- \\
% \hline

%  & None & -- & 33.10 & -- & -- & 40.3 & -- & 31.35 & -- & -- & 40.9 \\
%  & FPN & -- & 26.08 & -- & -- & 40.7 & -- & 25.31 & -- & -- & 41.3 \\
% \multirow{2}{*}{\textbf{DarkNet19}} & PANet & -- & 25.28 & -- & -- & 42.7 & -- & 24.54 & -- & -- & 43.4 \\
%  & BiFPNx1 & -- & -- & -- & -- & -- & -- & -- & -- & -- & -- \\
%  & BiFPNx2 & -- & -- & -- & -- & -- & -- & -- & -- & -- & -- \\
%  & BiFPNx3 & -- & -- & -- & -- & -- & -- & -- & -- & -- & -- \\
% \hline
\end{tabular}
\end{center}
% \vspace{-6pt}
\caption{Effectiveness of RFCR module}\label{tab:raw}
% \vspace{-12pt}
\end{table}

We now evaluate our raw feature collection and redistribution module, combined with various truncated backbones as well as feature aggregation paths adapted with our lightweight detection layers, in Table \ref{tab:raw}. Our method provides a consistent improvement in performance, irrespective of the backbone or the feature aggregation path that follows it. While this also comes at the cost of slight drop in FPS, overall the trade-off between the two is favorable for us, and thus the feature collection and redistribution module serves as a profitable lightweight addition to the model.

On taking a deeper dive into Table \ref{tab:raw}, we can notice that the effect of our feature redistribution is significantly more when there is no other feature aggregation method following it. This can be attributed to the fact that in the absence of any interaction between multi-scale features, except through the backbone itself, such a redistribution provides the much needed feature interaction. However, even with BiFPNx3, our method still gets a noticeable boost in performance, showing the importance of shortcut connections between non-adjacent layers.

Finally, we bring all methods discussed above together to do a combined component ablation study. The results are collected in Table \ref{tab:componenet}. We start with the MobileNetV2 ($\times$0.75) backbone for Jetson Nano, MobileNetV2 ($\times$1.4) backbone for Jetson Xavier NX and EfficientNet-B3 backbone for Jetson AGX Xavier, along with a PANet feature aggregation based YOLOv3 object detection head and lightweight detection layers.
Next, we test our RFCR module with and without truncating the backbone. While it is clear that RFCR module performs well in both scenarios, the drop in FPS for model with complete backbone is more as compared to the one with truncated backbone. This is due to the fact that the complete backbone has heavier layers towards the end, which makes the following feature aggregation layers heavier too.

As discussed in Section \ref{subsec:rfcr}, we also introduce additional 'shortcut' connections in our RFCR module independent of the detection head's output scales.
We notice that this additional 'shortcut' from shallower layers of the backbone further improves its accuracy, emphasizing the importance of low-level features for accurate detection tasks and the freedom our design provides in using more input features from the backbone than the number of output scales. Overall, we are able to both speed up the execution and improve accuracy by combining backbone truncation and RFCR module.

\subsection{Qualitative Analysis}
We also conduct a qualitative assessment of our RFCR module by visualizing various intermediate feature maps using our MobileNetV2x0.75 based model. We do a channel wise max pooling of the feature maps and then plot the obtained heatmaps scaled back to the original input image, as shown in Figure \ref{fig:qualitative}. For all 3 images, we notice that the raw feature maps (a) directly from the backbone are very noisy. While low-level raw features create a better boundary around the object, the attention patterns are discontinuous due to their smaller receptive field. On the other hand, high-level raw features create an inaccurate boundary which sometimes extends well beyond the object. However, just by doing a simple feature collection and redistribution, we obtain feature maps (b), in which we notice that various scales of features work together in harmony to obtain a better boundary of attention around the object.
We notice similar behavior between final feature maps (c) obtained right before the detection head, and final feature maps (d) obtained from a separately trained model which has the same architecture but without RFCR module, even though feature maps (d) are obtained after feature aggregation through PANet, which demonstrates that the RFCR module played an important role in filtering out the noise from raw features.

\begin{table}
\scriptsize
% \footnotesize
\begin{center}
\setlength\tabcolsep{4.5pt}
\begin{tabular}{c|ccc|c|c}
\toprule[2pt]
\textbf{Baseline} & \textbf{+Truncate} & \textbf{+RFCR} & \textbf{+Shortcut} & $\bm{AP^{50}}$ & \textbf{FPS} \\
\hline
\hline

% \multicolumn{6}{c}{\textit{MobileNetV2 x 0.75 on Jetson Nano}} \\
% \hline
 & \xmark & \xmark & \xmark & 68.67 & 28.16 \\
\cline{2-6}
\multirow{2}{*}{\textit{MobileNetV2 x 0.75}} & \cmark & \xmark & \xmark & 66.58 & 34.02 \\
\cline{2-6}
\multirow{2}{*}{\textit{on Jetson Nano}} & \xmark & \cmark & \xmark & 69.50 & 26.97 \\
\cline{2-6}
 & \cmark & \cmark & \xmark & 68.40 & 33.35 \\
\cline{2-6}
 & \cmark & \cmark & \cmark & 68.75 & 33.19 \\
\hline
\hline

% \multicolumn{6}{c}{\textit{MobileNetV2 x 1.4 on Jetson Xavier NX}} \\
% \hline
 & \xmark & \xmark & \xmark & 69.88 & 62.58 \\
\cline{2-6}
\multirow{2}{*}{\textit{MobileNetV2 x 1.4}} & \cmark & \xmark & \xmark & 69.67 & 67.38 \\
\cline{2-6}
\multirow{2}{*}{\textit{on Jetson NX}} & \xmark & \cmark & \xmark & 70.56 & 62.11 \\
\cline{2-6}
 & \cmark & \cmark & \xmark & 70.21 & 65.91 \\
\cline{2-6}
 & \cmark & \cmark & \cmark & 70.35 & 65.37 \\
\hline
\hline

% \multicolumn{6}{c}{\textit{EfficientNet-B3 on Jetson AGX Xavier}} \\
% \hline
 & \xmark & \xmark & \xmark & 72.05 & 73.15 \\
\cline{2-6}
\multirow{2}{*}{\textit{EfficientNet-B3}} & \cmark & \xmark & \xmark & 72.28 & 78.61 \\
\cline{2-6}
\multirow{2}{*}{\textit{on Jetson AGX}} & \xmark & \cmark & \xmark & 72.37 & 72.90 \\
\cline{2-6}
 & \cmark & \cmark & \xmark & 72.58 & 75.72 \\
\cline{2-6}
 & \cmark & \cmark & \cmark & 72.96 & 75.61 \\

\bottomrule[2pt]

\end{tabular}
\end{center}
% \vspace{-6pt}
\caption{Ablation study}\label{tab:componenet}
% \vspace{-10pt}
\end{table}

\subsection{Comparison with State-of-the-art Models}

\begin{table}
\scriptsize
% \footnotesize
\begin{center}
\setlength\tabcolsep{3pt}
\begin{tabular}{c|c|c|ccc|cc}
\toprule[2pt]
\multirow{2}{*}{\textbf{Model}} & \textbf{Input} & \textbf{Size} & \multicolumn{3}{|c|}{\textbf{FPS}} &
 \multicolumn{2}{|c}{$\bm{AP^{50}}$} \\
% \cline{4-8}
 & \textbf{Res.} & \textbf{(MB)} & \textbf{Nano} & \textbf{NX} & \textbf{AGX} & \textbf{VOC} & \textbf{COCO} \\
\hline
\hline

Tiny-YOLOv3 \cite{redmon2018yolov3} & 416 & 34.9 & 27.36 & 66.55 & 91.71 & 61.30 & 33.10 \\
% MobileNet-SSD \cite{huang2017speed} & 300 x 300 & 32 & 23.3 & -- & -- & 72.40 & -- \\
Tinier-YOLO \cite{fang2019tinier} & 416 & 8.9 & 30.14 & 68.73 & 92.09 & 65.70 & \textbf{34.00} \\
YOLO Nano \cite{wong2019yolo} & 416 & 4.0 & 13.62 & 54.03$^\ddagger$ & 85.81$^\ddagger$ & 69.10 & -- \\
YOLO-Fastest \cite{yolofastest} & 320 & 1.3 & \textbf{42.41} & \textbf{76.13} & \textbf{126.82} & 61.02 & -- \\
YOLO-Fastest XL \cite{yolofastest} & 320 & 3.5 & 27.93 & 61.33 & 108.76 & \textbf{69.43} & 32.45 \\
\hline
 & 416 & 5.2 & 19.87 & 58.24 & 71.16 & \textbf{72.39} & \textbf{36.44} \\
YOLO-ReT-M0.75 & 320 & 5.2 & 33.19 & 71.64 & 95.97 & 68.75 & 34.91 \\
 & 224 & 5.2 & \textbf{55.16} & \textbf{84.10} & \textbf{134.87} & 60.77 & 30.76 \\
\hline
 & 416 & 12.3 & 13.17 & 46.07 & 66.23 & \textbf{73.32} & \textbf{36.52} \\
YOLO-ReT-M1.4 & 320 & 12.3 & 23.01 & 65.37 & 93.49 & 70.35 & 35.77 \\
 & 224 & 12.3 & \textbf{43.16} & \textbf{84.32} & \textbf{113.94} & 62.91 & 31.63 \\
\hline
 & 416 & 28.3 & 6.35 & 28.83 & 49.07 & \textbf{76.49} & \textbf{39.12} \\
YOLO-ReT-EB3 & 320 & 28.3 & 10.96 & 44.59 & 75.61 & 72.96 & 36.51 \\
 & 224 & 28.3 & \textbf{18.57} & \textbf{54.87} & \textbf{93.55} & 65.52 & 33.11 \\
\bottomrule[2pt]

% YOLO-RED-EfficientNet-B0 & 416 & 17.4 & 11.54 & -- & -- & -- & -- \\
% YOLO-RED-EfficientNet-B0 & 320 & 17.4 & 17.82 & -- & -- & -- & -- \\
% YOLO-RED-EfficientNet-B0 & 224 & 17.4 & 33.58 & -- & -- & -- & -- \\
% YOLO-RED-DarkNet19 & 416 & 43.4 & 15.37 & -- & -- & -- & -- \\
% YOLO-RED-DarkNet19 & 320 & 43.4 & 24.54 & -- & -- & -- & -- \\
% YOLO-RED-DarkNet19 & 224 & 43.4 & 48.60 & -- & -- & -- & -- \\

\end{tabular}
\footnotesize{ $^\ddagger$ Calculated with INT8 optimization}
\end{center}
% \vspace{-10pt}
\vspace{-4pt}
\caption{Comparison with other state-of-the-art models}\label{tab:others}
% \vspace{-12pt}
\end{table}

We build models based on selected backbones using truncation and the RFCR module, and then compare them with 3 different input image resolutions to various state-of-the-art real-time object detection models with the same settings, as shown in Table \ref{tab:others}.
As expected, smaller input resolution results in a faster but less accurate detection model.
We also provide detailed evaluation on COCO dataset in Table \ref{tab:coco}, however the scope of comparison is limited as not all state-of-the-art models provide such detailed results.

\begin{table}
\scriptsize
% \footnotesize
\begin{center}
\setlength\tabcolsep{3pt}
\begin{tabular}{c|c|cccccc}
\toprule[2pt]
\multirow{2}{*}{\textbf{Model}} & \textbf{Input} & \multicolumn{6}{|c}{\textbf{COCO}} \\
% \cline{2-7}
 & \textbf{Res.} & $\bm{AP}$ & $\bm{AP^{50}}$ & $\bm{AP^{75}}$ & $\bm{AP^{s}}$ & $\bm{AP^{m}}$ & $\bm{AP^{l}}$ \\
\hline
\hline

Tiny-YOLOv3 \cite{redmon2018yolov3} & 416 & 15.3 & 33.1 & 12.4 & 4.4 & 15.2 & 25.1 \\
Tinier-YOLO \cite{fang2019tinier} & 416 & 17.0 & 34.0 & 15.7 & 4.8 & 17.3 & 26.8 \\
YOLO-ReT-M0.75 & 320 & 18.4 & 34.9 & 17.3 & 5.4 & 18.7 & 28.9 \\
YOLO-ReT-M1.4 & 320 & 19.1 & 35.8 & 18.4 & 5.8 & 19.6 & 30.2 \\
YOLO-ReT-EB3 & 320 & \textbf{19.7} & \textbf{36.5} & \textbf{19.3} & \textbf{6.3} & \textbf{20.3} & \textbf{31.1} \\

\bottomrule[2pt]
\end{tabular}
\end{center}
% \vspace{-6pt}
\caption{Evaluation on COCO dataset}\label{tab:coco}
% \vspace{-12pt}
\end{table}

When comparing on Jetson Nano, we find that our model YOLO-ReT-M0.75 at 320x320 resolution outperforms Tinier-YOLO by 3.05 mAP on Pascal VOC and 0.91 mAP on COCO, while executing faster by 3.05 FPS. On Jetson Xavier NX, our model YOLO-ReT-M1.4 at 320x320 resolution outperforms YOLO-Fastest-XL by 0.92 mAP on Pascal VOC and 3.34 mAP on COCO, while executing faster by 4.02 FPS.
Even though our YOLO-ReT-EB3 model at 416x416 resolution is able to push for the best performance while still executing real-time on Jetson Xavier AGX, it should be noted that at similar FPS, MobileNetV2 based models outperform EfficientNet based models.

\subsubsection{Comparison with YOLOv4-tiny}
\label{sec:yolov4}

The state-of-the-art in object detection is being pushed constantly, with multiple parallel works being published at any moment. YOLOv4-tiny \cite{wang2021scaled} has done the same for real-time object detection on edge devices, using novel training methods as well backbone scaling and customization for an improved object detection model that provides significantly better performance than any existing SOTA. While this work was done in parallel with ours, the improvements are undeniably significant to be ignored. Thus, we directly introduce our RFCR module into YOLOv4-tiny and YOLOv4-tiny (3l) without backbone truncation as these models are designed specifically for object detection and are trained from scratch without using transfer learning \cite{wang2021scaled}. We still use 4 inputs to our feature collection, and redistribute them based on the number of output scales present in the model, i.e. 2 for YOLOv4-tiny and 3 for YOLOv4-tiny (3l). 
We execute both models on Jetson Nano device since it has the least resource compared to the other two platforms. The final results are collected in Table \ref{tab:yolov4tiny}.

\begin{table}
\scriptsize
% \footnotesize
\begin{center}
\begin{tabular}{l|c|c|ccc}
\toprule[2pt]
\multirow{2}{*}{\textbf{Model}} & \textbf{Input} & \textbf{FPS} &
 \multicolumn{3}{|c}{\textbf{COCO}} \\
% \cline{4-8}
 & \textbf{Res.} & \textbf{Nano} & $\bm{AP}$ & $\bm{AP^{50}}$ & $\bm{AP^{75}}$ \\
\hline
\hline

YOLOv4-tiny & 416 & 29.55 & 21.7 & 40.2 & 22.5 \\
YOLOv4-tiny+RFCR & 416 & 27.81 & \textbf{22.9} & \textbf{41.5} & \textbf{23.3} \\
\hline
YOLOv4-tiny (3l) & 608 & 24.87 & 28.7 & 47.2 & 29.7 \\
YOLOv4-tiny (3l)+ RFCR & 608 & 21.40 & \textbf{29.3} & \textbf{48.1} & \textbf{30.5} \\

\bottomrule[2pt]

\end{tabular}
\end{center}
% \vspace{-10pt}
\caption{Comparison with YOLOv4-tiny \cite{wang2021scaled}}\label{tab:yolov4tiny}
% \vspace{-12pt}
\end{table}

We find that the RFCR versions of both these models are able to outperform their counterparts in terms of accuracy. Looking deeper, we find that our RFCR module provides larger improvement for YOLOv4-tiny, as compared to YOLOv4-tiny (3l). This is expected, as YOLOv4-tiny only has 2 output scales, which further necessitates multi-scale feature interaction, and helps us demonstrate the compatibility of our module with various detection models.
\section{Conclusion and Future Work}
\label{sec:conclusion}

This paper presents a novel raw feature collection and redistribution (RFCR) module, and a truncated backbone for improved transfer learning for object detection. These techniques complement each other, leading to both improved accuracy and efficiency for various lightweight architectures.
We believe that machine learning model designs targeting edge GPU devices have opened up a new avenue of research for edge computing, and can lead to a wide range of application possibilities.
Thus, further research on device specific model optimizations and neural architecture search can help push the technology forward effectively for real-time performing models.
Meanwhile, an in-depth understanding of interactions between various features for object detection can further enhance the information flow and is an important aspect of improving the model accuracy while maintaining its efficiency. In future, we aim to extend our RFCR module to other vision domains and a larger variety of models.
The source code of our design is opened to public at \url{https://github.com/prakharg24/yoloret}.
\section*{Acknowledgement}

This publication was made possible by grant AICC03-0324-200005 from Qatar National Research Fund (a member of Qatar Foundation), and grant MRC-05-110 from Hamad Medical Corporation. 
It is also partially supported by the National Research Foundation, Prime Minister's Office, Singapore under its Campus for Research Excellence and Technological Enterprise (CREATE) programme.
The findings herein reflect the work, and are solely the responsibility, of the authors.

{\small
\bibliographystyle{ieee_fullname}
\bibliography{egbib}
}

\end{document}